\documentclass[conference]{IEEEtran}
\IEEEoverridecommandlockouts
\usepackage{cite}
\usepackage{amsmath,amssymb,amsfonts}
\usepackage{algorithmic}
\usepackage{graphicx}
\usepackage{textcomp}
\usepackage{xcolor}
\def\BibTeX{{\rm B\kern-.05em{\sc i\kern-.025em b}\kern-.08em
    T\kern-.1667em\lower.7ex\hbox{E}\kern-.125emX}}
\begin{document}

\title{HistoTransfer: Understanding Transfer Learning for Histopathology \\
\thanks{The work was supported by NIDDK/NIH of the National Institutes of Health under award number K23DK117061-01A1 (Syed) and iTHRIV NIH-NCATS award UL1TR003015.} 
\thanks{* Co-Corresponding Author}
}


\author{
    \IEEEauthorblockN{Yash Sharma\IEEEauthorrefmark{2}, Lubaina Ehsan\IEEEauthorrefmark{2}, Sana Syed\IEEEauthorrefmark{1}\IEEEauthorrefmark{2}\IEEEauthorrefmark{3}, Donald E. Brown\IEEEauthorrefmark{1}\IEEEauthorrefmark{3}\IEEEauthorrefmark{4}}
    \IEEEauthorblockA{\IEEEauthorrefmark{2}Department of Pediatrics, University of Virginia}
    \IEEEauthorblockA{\IEEEauthorrefmark{3}School of Data Science, University of Virginia}
    \IEEEauthorblockA{\IEEEauthorrefmark{4}Department of Engineering Systems and Environment, University of Virginia
    \\\{ys5hd, le7jg, sana.syed, deb\}@virginia.edu}
}

\maketitle

\begin{abstract}
Advancement in digital pathology and artificial intelligence has enabled deep learning-based computer vision techniques for automated disease diagnosis and prognosis. However, WSIs present unique computational and algorithmic challenges. WSIs are gigapixel-sized, making them infeasible to be used directly for training deep neural networks. Hence, for modeling, a two-stage approach is adopted: Patch representations are extracted first, followed by the aggregation for WSI prediction. These approaches require detailed pixel-level annotations for training the patch encoder. However, obtaining these annotations is time-consuming and tedious for medical experts. Transfer learning is used to address this gap and deep learning architectures pre-trained on ImageNet are used for generating patch-level representation. Even though ImageNet differs significantly from histopathology data, pre-trained networks have been shown to perform impressively on histopathology data. Also, progress in self-supervised and multi-task learning coupled with the release of multiple histopathology data has led to the release of histopathology-specific networks. In this work, we compare the performance of features extracted from networks trained on ImageNet and histopathology data. We use an attention pooling network over these extracted features for slide-level aggregation. We investigate if features learned using more complex networks lead to gain in performance. We use a simple top-k sampling approach for fine-tuning framework and study the representation similarity between frozen and fine-tuned networks using Centered Kernel Alignment. Further, to examine if intermediate block representation is better suited for feature extraction and ImageNet architectures are unnecessarily large for histopathology, we truncate the blocks of ResNet18 and DenseNet121 and examine the performance.
\end{abstract}


\section{Introduction}

Histopathology comprises an essential step in the diagnosis, prognosis, and treatment response of patients with cancer and gastrointestinal diseases, among others \cite{esteva2021deep}. In recent years, digital pathology has seen an increase in the availability of digitized whole slide images (WSIs) and consequently in the development of novel computational frameworks for computer-aided diagnosis. However, this area poses its own unique challenges, including variability in histopathological features across diseases, limited data for training, and the extremely high resolution of WSI images ($\sim100k \times 100k$ pixel). Large WSI sizes make them computationally infeasible to be directly used for the training of deep learning-based techniques. Downsampling images for training leads to loss of relevant cellular and structural details pertinent for diagnosis. Therefore, several recently proposed frameworks follow a two-stage modeling approach where patch-level feature extraction is performed followed by an independent aggregation approach for obtaining WSI prediction. 


\begin{figure}[t]
\centerline{\includegraphics[width=\linewidth]{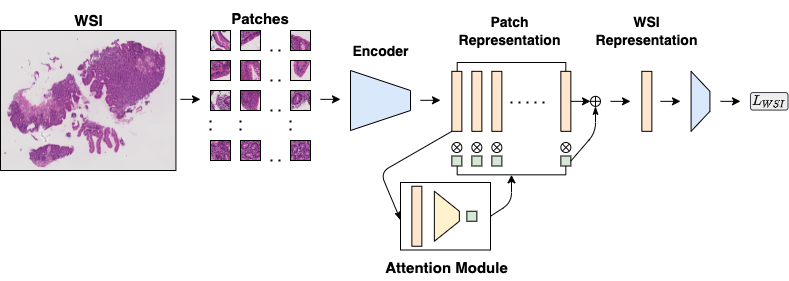}}
\caption{Histopathology Model Architecutre}
\label{fig:arch}
\end{figure}

Generally, approaches use a patch encoder followed by either a machine learning model or mathematical aggregation such as mean or max pooling for slide-level prediction. The patch encoder can be trained in both supervised or unsupervised settings. With the supervised training set-up, patch-level labels are required for training. Most proposed approaches utilize a Multiple-Instance Learning (MIL) based set-up to train on noisy labels under the mathematical assumption that all the patches from a positive WSI are positive, which may not be true. Alternatively, these approaches require pathologists to annotate the complete slide at the cellular level for patch-level labels. Another commonly used approach employs unsupervised methods such as autoencoder for learning patch representation. However, these methods do not guarantee that discriminative features for normal and diseased tissues are being learned \cite{srinidhi2020deep}.
These limitations have increased the interest in an end-to-end training framework using WSIs \cite{chikontwe2020multiple, xie2020beyond, sharma2021cluster}. 

In this paper, we focus on a transfer learning based approach (Fig \ref{fig:arch}). Further, we extend the training by fine-tuning our encoder in an end-to-end framework. 
We use a top-k sampling approach for fine-tuning the network learned in the transfer learning stage. Prior approaches have demonstrated that fine-tuning the base architecture in conjunction with the aggregation modules can lead to better-optimized models \cite{sharma2021cluster}. In contrast, the framework recently proposed in \cite{lu2021data} aggregates the representation generated by ImageNet pre-trained network and has shown to perform competitively with other approaches. This raises 3 primary questions which we address in this work:  
\begin{enumerate}
    \item How much does the selection of pre-trained networks impact the performance of the framework? We experiment with ResNet, DenseNet, and EfficientNet architectures pre-trained on ImageNet data. Moreover, we use publicly available self-supervised learning and multi-task learning models trained on multiple histopathology data to compare if domain-specific encoder improves performance.
    \item Can we fine-tune the patch encoder further and consequently improve the performance of the end-to-end framework? 
    \item Raghu et al.\cite{NEURIPS2019_eb1e7832} and Ke et al. \cite{ke2021chextransfer} have demonstrated that popularly used architectures developed for 1000-class ImageNet data can be pruned or replaced with a smaller model for binary or multi-class medical imaging problems. Also, it is well understood that, in the deeper networks, as we reach the classification layer, the representation becomes more domain-specific. To investigate this and examine if later blocks of ImageNet architecture can be removed without affecting the performance, we extract intermediate block representation from ResNet18 and DenseNet121 and use it with our transfer learning architecture for the classification task. 
\end{enumerate}

\section{Related Work}

\subsection{Histopathology Pre-Trained Encoder}
Ciga et al. \cite{ciga2020self} used a contrastive self-supervised learning method for training on 60 histopathology dataset images to develop a domain-specific encoder. They demonstrated that linear classifiers trained on top of their encoder features perform better than ImageNet pre-trained networks, boosting task performance up to 7.5\% in accuracy on out-of-domain histopathology datasets. 
We use their publicly released ResNet18 weights in our transfer learning experiments for comparison. Mormont et al. \cite{mormont2020multi} used multi-task learning as a way of pre-training models for developing histopathology-specific classification models. They used data from over 22 publicly available histopathology classification tasks with over 81 classes and 900k images for training their model. They reached similar conclusions as observed in the \cite{ciga2020self}, domain-specific pre-trained encoder benefits max in feature extraction tasks. However, fine-tuning can recover the lack of specificity of the ImageNet pre-trained encoder and gives comparable performance to the domain-specific encoder. We use their publicly released weights of ResNet50 and DenseNet121 for comparison.

\subsection{Transfer Learning}
Ke et al. \cite{ke2021chextransfer} experimented with different ImageNet pre-trained architecture on Chest X-ray classification task. They found no correlation between the ImageNet performance and chest x-ray performance, concluding that improvements observed on ImageNet may not be directly transferable to medical imaging tasks. However, they found a significant boost in performance associated with ImageNet pre-training against models without any pre-training. Further, they demonstrated that final blocks of pre-trained models could be truncated without observing a statistically significant drop in performance. Owing to the unique problems of big size in histopathology data compared to other medical imaging problems, we extend their work with our transfer learning approach to study the relationship between ImageNet and Histopathology performance. On retinal fundus and chest x-ray images, Raghu et al. \cite{NEURIPS2019_eb1e7832} evaluated the performance of smaller simpler convolutional architectures against standard ImageNet models concluding that smaller models perform comparably to bigger models, and ImageNet performance is not indicative of medical performance. Using SVCCA for computing representation similarity, they investigated hidden representation similarity between randomly initialized, pre-trained, and fine-tuned encoder concluding that pre-training does affect the hidden representation and standard ImageNet models do not change significantly through the fine-tuning process. 

\section{Methods}

\subsection{Background}
For digital pathology classification problems, WSIs ($W$) of patients are available along with their disease labels. Typically, a WSI is in dimensions ranging from $50k\times50k$ to $100k\times100k$ pixels, making it computationally infeasible to use them for training directly. Hence, using the Otsu thresholding approach and sliding window approach, patches containing substantial tissue area ($>50\%$) of desirable size are extracted. 
As we approach this problem with the MIL set-up, we assume at least one of the patches extracted from the diseased slide contains diseased features, while a normal slide has all healthy patches.

For experimentation, we divide our training into 2 stages: (1) Transfer Learning Training (2) Fine-tuning. 


\subsection{Transfer Learning Architecture}
We use a pre-trained encoder for extracting patch representation followed by the weighted-average aggregation approach proposed in Ilse et al.\cite{ilse2018attention} for aggregating the patch-level representation to WSI-level representation (Fig \ref{fig:arch}). We use a two-layered neural network to compute weights for each patch in the WSI for aggregation modules. Using a frozen pre-trained encoder allows us to use all patch representation for training attention aggregation modules without any computational limitations. In each training epoch, the aggregation module is trained for generating differentiable WSI representation. 

\subsection{Fine-tuning Architecture}
We can not use all the patches for fine-tuning the encoder in an end-to-end learning framework due to computational limitations. Therefore, we extend the model developed in the transfer learning step by sampling top-k attended patches based on the weights assigned by the attention module. Using top-k sampled patches, we fine-tune the architecture along with the encoder trained in the previous step. Further, we investigate the similarity between hidden block representation in frozen and fine-tuned models using centered kernel alignment (CKA) to study where the most update is happening \cite{kornblith2019similarity}. 

\begin{table}[t]
\caption{Comparison of different architectures Gastrointestinal and Camelyon task AUC against ImageNet Top 1 Accuracy}
\begin{center}
\begin{tabular}{|c|c|c|c|}
\hline
\textbf{Model}&\textbf{Gastro AUC}&\textbf{Camelyon AUC}&\textbf{ImageNet Acc@1} \\
\hline
ResNet18 & $90.3$ & $76.5$ & $69.8$ \\
\hline
ResNet34 & $90.8$ & $71.9$ & $73.3$ \\
\hline
ResNet50 & $88.4$ & $71.9$ & $76.1$ \\
\hline
DenseNet121 & $91.1$ & $79.9$ & $74.4$ \\
\hline
DenseNet169 & $90.8$ & $75.0$ & $75.6$ \\
\hline
EfficientNetB0 & $86.6$ & $72.5$ & $76.3$ \\
\hline
EfficientNetB1 & $90.0$ & $76.9$ & $78.8$ \\
\hline
EfficientNetB2 & $87.8$ & $69.3$ & $79.8$ \\
\hline
EfficientNetB3 & $90.5$ & $69.9$ & $81.1$ \\
\hline
\end{tabular}
\label{tab:ImageNetvsHisto}
\end{center}
\end{table}

\section{Experiments and Results}

\subsection{Data and Model Description}

We performed our experiments on two binary classification datasets - a gastrointestinal dataset and a breast cancer dataset. The gastrointestinal dataset contains $413$ high-resolution WSIs obtained from digitizing $124$ H\&E stained duodenal biopsy slides at $40\times$ magnification. These biopsies were from children who went endoscopy procedures at the University of Virginia Hospital. There were $63$ children with Celiac Disease (CD) and $61$ healthy children (with histologically normal biopsies). A $65$\%-$15$\%-$20$\% split was used to split data for training, validation, and testing. Patches with at least $50$\% tissue area of size $512 \times 512$ were extracted from each WSI. The other dataset used in our study was the publicly available CAMELYON16 dataset \cite{bejnordi2017diagnostic} for breast cancer metastasis detection containing $270$ train WSIs and $129$ test WSIs. Patches of size $512 \times 512$ at $20\times$ magnification were extracted. 

\subsection{Transfer Learning Experiment}

Using our transfer learning architecture, we investigated if the performance gains observed with ResNet, DenseNet, and EfficientNet architectures on ImageNet are transferable to Histopathology data. Table \ref{tab:ImageNetvsHisto} demonstrates that there is no relation between histopathology AUCs and ImageNet top-1 accuracy. We observed that within each architecture family, smaller architectures achieve similar, if not better, performance compared to bigger architectures. Thus, increasing complexity within the family does not yield an increase in histopathology performance. Among all the architectures, DenseNet121 performed the best. On average, ResNet and DenseNet family of algorithms perform better than EfficientNet. This corroborates the observation made in CheXtransfer that recently released algorithms are generated through neural architecture search optimizing performance on ImageNet, but such performance gain may not be directly transferable to medical image datasets.

\begin{table}[htbp]
\caption{Comparison of ImageNet PreTrained Encoder against Histopathology Specific Enocoder}
\begin{center}
\begin{tabular}{|c|c|c|}
\hline
\textbf{Model}&\textbf{Training Strategy}&\textbf{Gastro AUC} \\
\hline
ResNet18 & ImageNet Training & $90.3$ \\
\hline
ResNet18 & Histopathology Self-Supervised Learning & $93.7$\\
\hline
DenseNet121 & ImageNet Training & $91.1$ \\
\hline
DenseNet121 & Histopathology Multi-task Learning & $93.1$ \\
\hline
ResNet50 & ImageNet Training & $88.4$ \\
\hline
ResNet50 & Histopathology Multi-task Learning & $90.6$ \\
\hline
\end{tabular}
\label{tab:DomainSpecific}
\end{center}
\end{table}

\subsection{Histopathology Encoder vs ImageNet Encoder}

For evaluating encoder trained on histopathology, we only used the gastrointestinal dataset as the Camelyon dataset was part of the training set in both the references \cite{mormont2020multi, ciga2020self}. We observed that encoders trained using both self-supervised learning and multi-task learning on histopathology data lead to superior performance resulting in higher AUC in our transfer learning architecture. This satisfies the observation from \cite{mormont2020multi} that domain-specific encoders perform better for feature extraction than ImageNet encoders. And we suggest that these publicly available weights should be used for weight initializations in histopathology-specific tasks. Further, to check if by fine-tuning the ImageNet encoder, we can recover the lack of specificity of ImageNet features, we fine-tune DenseNet121 and ResNet18 initialized using ImageNet. (Table \ref{tab:DomainSpecific})

\begin{table}[htbp]
\caption{Fine-Tuning Experiment}
\begin{center}
\begin{tabular}{|c|c|c|c|}
\hline
\textbf{Data}&\textbf{Model}&\textbf{Frozen AUC}&\textbf{FineTune AUC} \\
\hline
Gastro & ResNet18 & $90.3$ & $91.3$ \\
\hline
Camelyon & ResNet18 & $76.5$ & $84.5$ \\
\hline
Gastro & DesneNet121 & $91.1$ & $92.9$ \\
\hline
Camelyon & DenseNet121 & $79.9$ & $89.0$ \\
\hline
\end{tabular}
\label{tab:FineTune}
\end{center}
\end{table}

\subsection{Finetuning Experiment}

We use the aggregation module developed in transfer learning architecture and extract the top 64 patches for each WSI for fine-tuning the base encoder without computational limitations. Using the transfer learning architecture as the initialization, we train the whole architecture end-to-end while updating the parameters of our base encoder with a learning rate of $1e-5$. As shown in Table \ref{tab:FineTune}, we significantly improved the AUC on the Camelyon dataset while achieving minor improvements in the gastrointestinal dataset. For the gastrointestinal dataset, we could not recover the difference observed between ImageNet initialization and multi-task learning initialization. We hypothesize this is happening because domain-specific encoders are superior for histopathology classification tasks, and we require either patch-level annotations or a larger WSI dataset for recovering the difference.
Further, using CKA, we study the block-wise representation of the encoders to understand the fine-tuning step. Both ResNet18 and DenseNet121 consist of 4 blocks; we extract representation at each of these blocks and compute the similarity between frozen encoder representation and fine-tuned encoder representation. 

\begin{figure}[t]
\centerline{\includegraphics[width=0.8\linewidth]{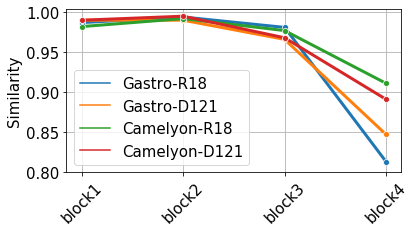}}
\caption{Intermediate block representation similarity between frozen and fine-tuned encoder computed using CKA. R18 is ResNet18 and D121 is DenseNet121.}
\label{fig:finetune}
\end{figure}

\begin{table}[htbp]
\caption{Truncated Model Experiment}
\begin{center}
\begin{tabular}{|c|c|c|c|}
\hline
\textbf{Model}&\textbf{Gastro AUC}&\textbf{Camelyon AUC} \\
\hline
ResNet18 & $90.3$ & $76.5$ \\
\hline
ResNet18-Minus1 & $90.8$ & $78.5$ \\
\hline
ResNet18-Minus2 & $88.3$ & $70.5$ \\
\hline
ResNet18-Minus3 & $84.5$ & $39.1$ \\
\hline
DenseNet121 & $91.1$ & $79.9$ \\
\hline
DenseNet121-Minus1 & $91.8$ & $80.8$ \\
\hline
DenseNet121-Minus2 & $91.7$ & $76.7$ \\
\hline
DenseNet121-Minus3 & $85.4$ & $46.5$ \\
\hline
\end{tabular}
\label{tab:truncate}
\end{center}
\end{table}

\subsection{Truncated Architecture}

We observed that relatively maximum change in representation is happening in the later blocks (Fig \ref{fig:finetune}), aligning with the understanding that initial layers of deep architectures learn more general representation, and as we move towards penultimate layers, more task-specific patterns are learned. Motivated by these findings and conclusions from the \cite{ke2021chextransfer} highlighting that bigger models can be pruned without significantly impacting the performance. We examined if blocks of these deeper architectures can be truncated without affecting the performance. We pruned blocks of ResNet18 and DenseNet121 and reported the performance using transfer learning architecture (Table \ref{tab:truncate}). We found that removing the final block (minus1) in both the models does not impact our performance, and there is a small drop when we remove both blocks 3 and 4 (minus 2). This aligns with our observations from similarity analysis that later blocks in the deep models learn domain-specific features and can be pruned without significantly impacting the performance of the feature extraction step resulting in a parameter-efficient model and higher-resolution class activation map visualizations (CAM) for interpretability.

\section{Discussion}

In this paper, we conclude that gains observed in the ImageNet dataset using bigger models are not transferable to histopathology datasets. Smaller models like ResNet18 and DenseNet121 tend to perform better than their larger counterparts. We demonstrated that domain-specific encoders trained on multiple histopathology datasets result in superior features for classification tasks than their ImageNet trained counterparts and should be used for weight initialization in histopathology tasks. The performance can be further improved by fine-tuning the base encoder on top-attended patches. Using CKA, we systematically studied the representational similarity between frozen and fine-tuned encoders and performed block pruning, concluding that later blocks of our encoder can be removed for feature extraction without impacting the performance of the model leading to a parameter-efficient model and higher resolution CAMs.

\bibliographystyle{IEEEtran}
\bibliography{IEEEabrv}

\end{document}